\documentclass[letterpaper, 10 pt, conference]{ieeeconf}  

\IEEEoverridecommandlockouts                              

\title{\LARGE \bf
Human-Robot Teaming Field Deployments: A Comparison Between Verbal and Non-verbal Communication
}

\author{Tauhid Tanjim$^{1}$, Promise Ekpo$^{1}$, Huajie Cao$^{2}$ , Jonathan St. George $^{3}$, \\Kevin Ching$^{3}$, Hee Rin Lee$^{2}$, and Angelique Taylor$^{1}$
\thanks{*This material was supported by the National Science Foundation under Grant No. IIS-2423127.}
\thanks{$^{1}$T. Tanjim, P. Ekpo, and A. Taylor are with the Department of Information Science at Cornell University, Ithaca, NY 14850, USA {\tt\small\{tt485, poe6, amt298\}@cornell.edu}
        }%
\thanks{$^{2}$H. Cao and H. R. Lee are with the Department of Media and Information, Michigan State University, East Lansing, MI 48824, USA {\tt\small \{caohuaji, heerin\}@msu.edu}}        
\thanks{$^{3}$J. St. George and K. Ching is with Weill Cornell Medicine, Cornell University, New York, NY 10065, USA {\tt\small \{jos7007, kec9012\}@med.cornell.edu}}%
}

\usepackage{graphicx}
\usepackage{colortbl}
\usepackage{hyperref}
\usepackage{balance}
\usepackage{flushend}
\usepackage{dblfloatfix}
\usepackage{multirow}
\usepackage{threeparttable}
\usepackage{comment}

\usepackage[normalem]{ulem}
\begin{document}

\maketitle
\thispagestyle{empty}
\pagestyle{empty}

\begin{abstract}

Healthcare workers (HCWs) encounter challenges in hospitals, such as retrieving medical supplies quickly from crash carts, which could potentially result in medical errors and delays in patient care. 
Robotic crash carts (RCCs) have shown promise in assisting healthcare teams during medical tasks through guided object searches and task reminders.
Limited exploration has been done to determine what communication modalities are most effective and least disruptive to patient care in real-world settings. 
To address this gap, we conducted a between-subjects experiment comparing the RCC’s verbal and non-verbal communication of object search with a standard crash cart in resuscitation scenarios to understand the impact of robot communication on workload and attitudes toward using robots in the workplace. 
Our findings indicate that verbal communication significantly reduced mental demand and effort compared to visual cues and with a traditional crash cart. 
Although, frustration levels were slightly higher during collaborations with the robot compared to a traditional cart. 
These research insights provide valuable implications for human-robot teamwork in high-stakes environments.


\end{abstract}

\section{INTRODUCTION}

Hospital emergency rooms (ER) are fast-paced, high-stress environments where healthcare workers (HCWs) collaboratively treat patients \cite{gray2019workplace}.
Due to the time-sensitive and unpredictable nature of the ER environment, tasks are often executed in parallel \cite{bagnasco2013identifying, o2008professional}, increasing cognitive load. 
Hence, HCWs are prone to medical errors, and ongoing efforts have been made to identify ways to alleviate these errors.
Closed-loop communication  \cite{Diaz_Dawson_2020, Salik_2023a} helps mitigate communication-related errors and ensure the accuracy and acknowledgment of critical information shared among team members. 
However, even when HCWs use closed-loop communication, medical errors might stem from environmental noise from multiple people speaking simultaneously, and disruptions from distraught family members.
Additional support is needed to prevent errors when traditional protocols fail.

 
How can robots reduce medical errors committed by clinical teams in critical care settings?
A prominent source of medical errors stems from retrieving medical supplies from crash carts \cite{jacquet2018emergency}, which contain medical supplies and medications. 
The urgency of patient care often leads to impulsive shuffling through drawers, causing care delays, increased cognitive stress, and potential errors in object retrieval. HCWs often shuffle through cart drawers as the configuration may vary from one healthcare facility to another \cite{jacquet2018emergency,makkar2016study}. 
Furthermore, variations in adherence to stocking standardized inventory across healthcare facilities prevent easy access to supplies. Each drawer in the cart is divided into multiple sections containing different supplies, requiring clinicians to quickly locate the necessary items under intense pressure. 
As a result, searching for supplies could lead to patient care delays and increase HCWs’ cognitive load during emergencies \cite{jacquet2018emergency, ruggles2010standardizing}. 
Hence, HCWs could benefit from support in retrieving relevant items from crash carts. To address these challenges, embodying robotic assistance within crash carts is a promising direction in improving object retrieval, understanding the interactions of HCWs with the crash cart during ER procedures, and informing design strategies for robotic assistance without disrupting patient care. 
This research investigates how robotic crash carts (RCC) can retrieve supplies more efficiently.  
RCCs have the potential to improve the medical team's effectiveness through object search guidance. 

\begin{figure}[t]
    \centering
    \includegraphics[width=0.8\linewidth]{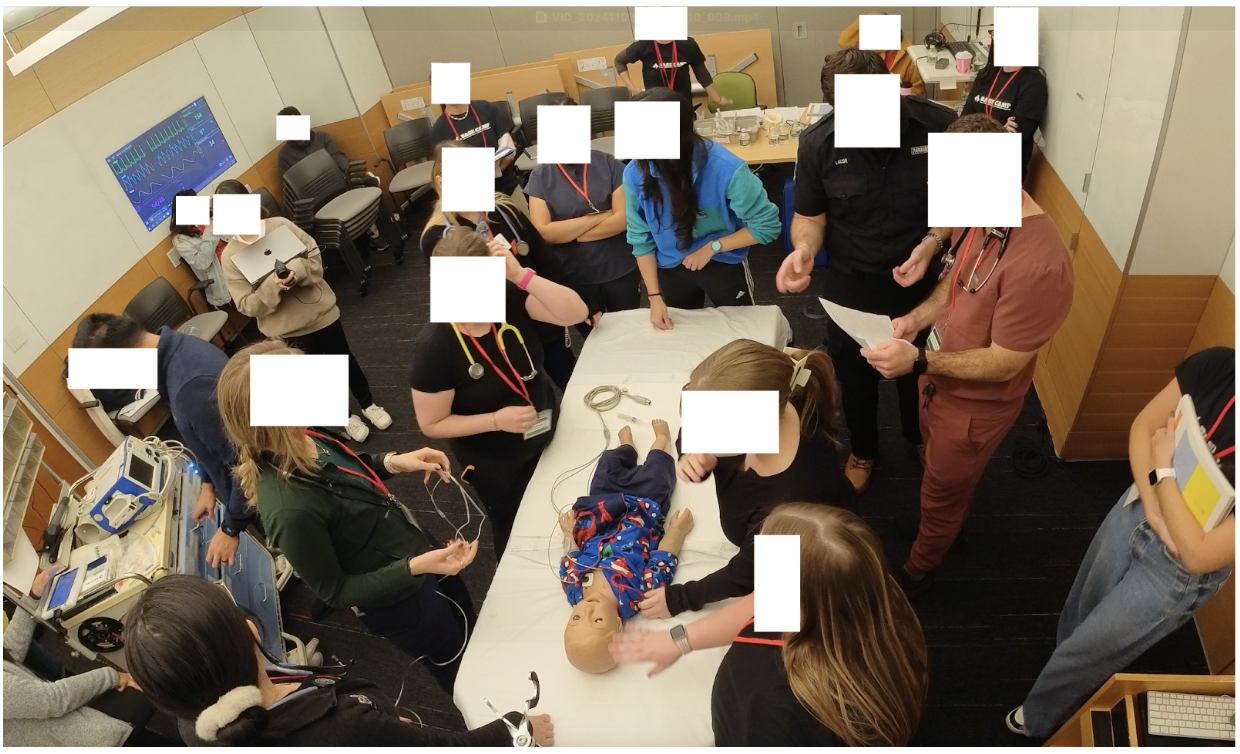} 
    \caption{Study session with participants engaging with a crash cart robot.}
    \label{fig:emergecy_room_setup}
\end{figure}


Prior work has investigated human-robot teaming in many safety-critical domains, including military operations \cite{jentsch2016human, barnes2019human, adams2024human}, disaster response \cite{kruijff2014designing}, and space exploration \cite{bordini2009formal, gervits2018shared}. 
Despite this progress, integrating robots into the ER remains a significant challenge because it is fast-paced and high-stakes, necessitating seamless and efficient human-robot communication. 
However, robots could introduce additional layers of complexity. 
When poorly aligned with clinical workflows, robots risk becoming a source of friction, adding to users' cognitive workload and potentially disrupting critical human-human communication.
Prior work has demonstrated the promise of using robots embodied in crash carts to intervene as objective actors in assisting ER teams during medical procedures since they are not susceptible to human error and psychological strain \cite{taylor2019coordinating, taylor2024towards, taylor2025rapidly}.


Despite these advancements, significant gaps remain in integrating RCCs into ER teams.    
First, a comprehensive investigation is needed to determine how RCCs can assist ER teams through guided object searches in real-world settings. 
Second, we need to explore the effectiveness and trade-offs of verbal and non-verbal communication within human-robot teams in the ER.   
Addressing these gaps are necessary for successful deployment of robots in ERs since poorly designed robotic communication could be counterproductive leading to dire consequences such as heightened fatigue of HCWs, patient safety risks, medical errors, and limited user adoption of robots in the workplace \cite{ rivera2010interruptions, sutherland2023fatigue, taylor2025rapidly, taylor2024towards}. 
Finally, field studies capture real-world behaviors and interactions compared to controlled, in-lab studies.


To address these gaps, we conducted an experimental field study to evaluate the effectiveness of robot assistance using verbal and non-verbal cues during human-robot teaming scenarios. 
The study was carried out in medical simulation training sessions.
We assess the trade-offs between two robot communication modes for minimal patient care interruption to determine the effectiveness of object search guidance using robotic speech and LED blinking (see Figure \ref{fig:robot_cart}).  
We measure HCWs’ workload and attitudes towards robots in medical training field deployments to gain insights for introducing robots as collaborative team members in these settings.
Furthermore, we address the following \textbf{research questions}: \textbf{R1:} How do RCCs impact team collaboration in high-stakes team collaborations in terms of workload compared to standard crash carts? \textbf{R2:} How do clinicians’ attitudes toward collaborative robotic assistance differ using speech-based and LED-based blinking in ER scenarios? 


The \textbf{contributions of this paper are fourfold}:
First, we present a comprehensive investigation of the RCC’s effectiveness during resuscitation to guide object search and provide insights for HRI researchers to integrate robots into safety-critical team collaborations.
Second, we conduct studies at a medical training event with a high degree of realism, strengthening the validity of our results.
Third, we compare RCC LED blinking and speech for communicating object guidance to a standard crash cart and demonstrate that speech guidance significantly reduces mental demand and effort. 
Fourth, our results indicate that HCWs' attitudes toward cooperative robot assistance in high-stakes team collaborations are positive regarding the use of LED blinking and speech for object search. 

Our study provides new knowledge about HCWs' needs and perspectives regarding robot assistants in high-stakes team collaborations, informing future design considerations for effective object-search guidance and task reminders.
Our study shows the potential of RCCs to enhance collaboration in terms of perceived workload and healthcare workers' attitudes toward robotic assistance in high-stakes environments.

\section{RELATED WORK}

\subsection{Human-Robot Teaming in High-Stakes Environments}

Human-robot teaming has gained significant attention in HRI in recent years, particularly in high-stakes environments such as search and rescue operations \cite{nourbakhsh2005human}, space missions \cite{ gervits2018shared}, disaster response \cite{kruijff2014designing}, and military operations \cite{barnes2019human, adams2024human}.
In the human-robot teaming literature, robots assist teams in decision-making to reduce workload, improve collaboration, and speed up task execution \cite{chen2014human, ma2022metrics}.
However, human-robot teaming in safety-critical healthcare environments remains underexplored \cite{taylor2025rapidly}.
The most closely related work has investigated human-robot teaming in the ER \cite{haripriyan2024human, jamshad2024taking, taylor2019coordinating}.  
These environments are complex, overcrowded, and require quick decision-making to operate effectively under time constraints with changing patient conditions \cite{taylor2025rapidly}. 
Specifically, there is limited understanding of how robots can effectively communicate with clinical teams without increasing their workload and how healthcare workers perceive and engage with these robots during fast-paced, real-world human-robot team interactions.
Thus, further studies are required to understand the effectiveness of robot communication strategies in reducing disruptions, human error, and workload and accounting for alarm fatigue.

\subsection{Multimodal Interaction in HRI}


Robots have long used multimodal interaction in the HRI field to facilitate fluent human-robot collaboration using verbal and non-verbal communication such as speech, gestures, visual cues (e.g., lights), and physical touch \cite{su2023recent, Bonarini_2020}. 
Prior research has explored these forms of communication in diverse fields such as aerial robotics \cite{szafir2015communicating}, agricultural robotics \cite{kaszuba2023speech, kamboj2022examining}, and healthcare robotics \cite{haripriyan2024human}. 
However, research on multimodal communication in safety-critical healthcare applications is limited.
Inspired by the research showing how robot verbal encouragement \cite{tang2024assisting} and nonverbal cues \cite{haripriyan2024human, breazeal2005effects} can improve group participation and support team dynamics, our work seeks to understand the impact of these cues on HCWs’ workload and willingness to adopt robots.

\subsection{HRI Field Studies}

HRI field studies offer critical knowledge about effective human-robot collaboration in real-world settings, such as deploying and observing trashcan robots in public spaces like city squares \cite{brown2024trash}.
These studies examine the ways in which humans and robots coordinate their actions, how the robots adapt to environmental change, and how to ensure trust, reliability, and safety between team members \cite{lee2023situating, alami2006safe}. 
Prior work focused on human-robot teaming in high-stakes environments, including simulated settings, such as escape-room scenarios mimicking hazard cleanup \cite{haripriyan2024human} and virtual testbeds designed to simulate emergency field studies \cite{raimondo2022trailblazing, wong2021remote}.  
However, deploying robots in high-stakes team settings can be challenging due to the limited availability of HCWs for research participation, logistical concerns, and ethical concerns regarding field studies.
Our work addresses this gap by deploying robots in high-stakes, high-fidelity simulation environments with stakeholders \cite{heinold2024advaned}.

\section{Crash Cart Robot Field Deployments}

\subsection{Robotic Crash Cart}

\begin{figure}[t]
    \centering
    \includegraphics[width=1.0\linewidth]{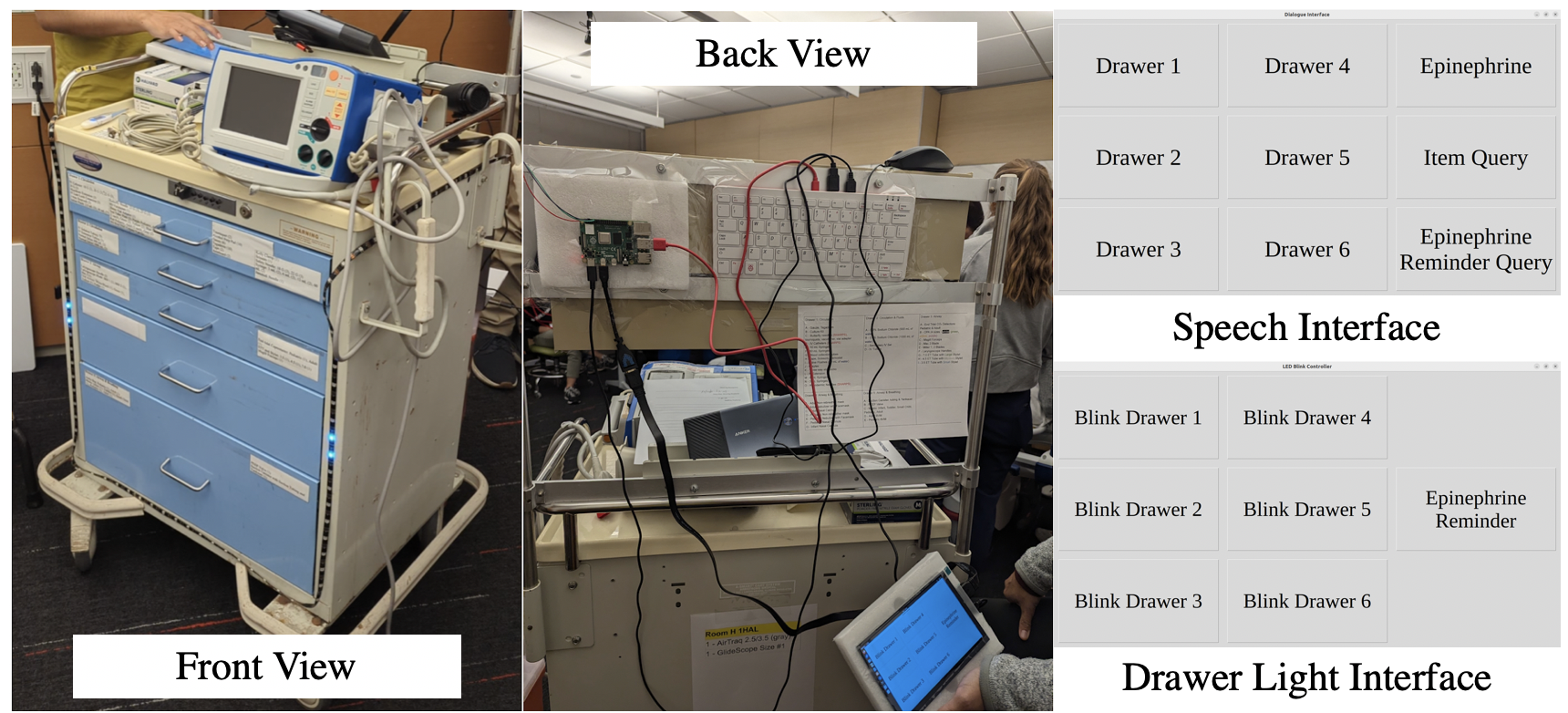} 
    \caption{The robotic crash cart is a Wizard-of-Oz platform, controlled by a Raspberry Pi connected to a tablet with a user interface that helps teams search for supplies using drawer lights and speech.}
    \label{fig:robot_cart}
\end{figure}

\begin{figure}[t] 
	\centering 
	\includegraphics[width=0.5\textwidth]{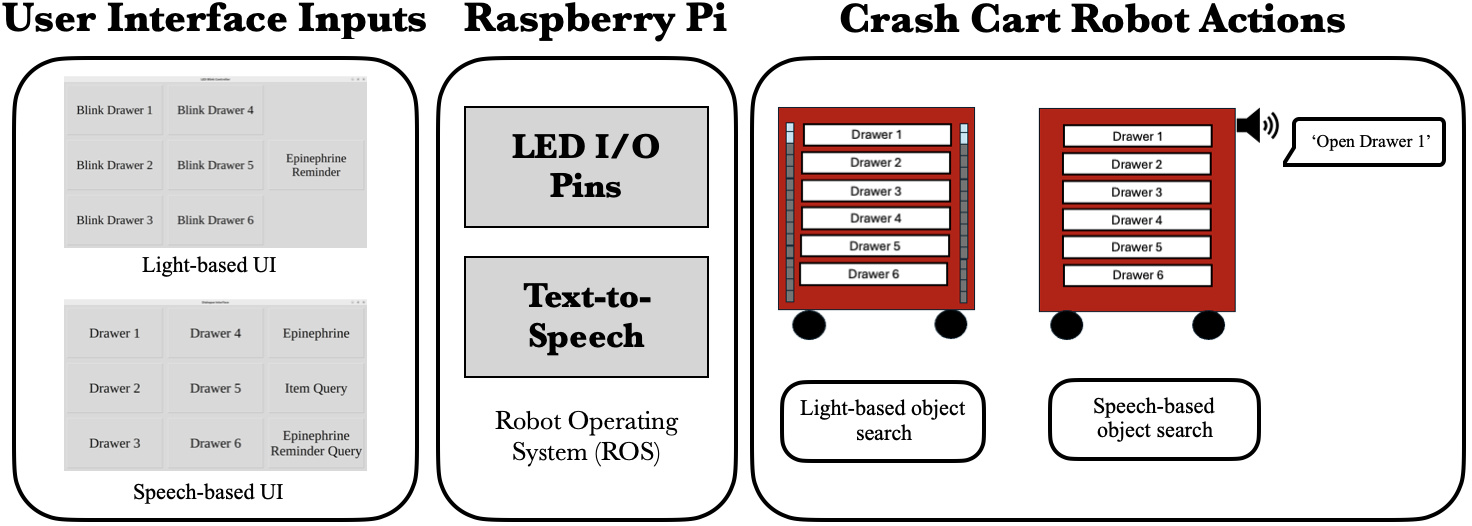} \caption{System diagram of the robotic crash cart illustrating the portable touch screen User Interface, processing via Raspberry Pi 4B using ROS, and outputs through light strips and audio signals.} 
	\label{fig:experiment_diagram} 
\end{figure}

We conducted experiments using the Crash Cart Robotic Toolkit\footnote{\url{https://github.com/Cornell-Tech-AIRLab/crash_cart_robot_tutorial} \label{github_link}} by Taylor et al. \cite{taylor2025rapidly}.
The RCC setup is made up of several input/output devices, including a touchscreen tablet (see Figures \ref{fig:robot_cart} and \ref{fig:experiment_diagram}), speaker, Raspberry Pi, and LED indicators. 
Buttons on the touch screen enable participants to teleoperate the robot’s speech and turn drawer LEDs on and off.  
For example, to activate speech-based object search, the teleoperator presses a button that prompts the robot to say ``Which item are you searching for?”
Then, the participant could respond by saying the item, e.g., “Epinephrine”. 
The wizard was positioned near the robot and looked up the item location from a catalog and pressed a button to enable the robot to say the drawer location, e.g ``Drawer 1''. 


\subsection{Study Design} 

We conducted an IRB-approved (IRB\#) experimental study during a two-day intensive interprofessional medical simulation-based event at a medical school in the global north to evaluate the use of speech and LED blinking for communicating object search guidance during Emergency Medicine simulation training sessions.
In this between-subjects study, participants were randomly assigned to engage in one of three conditions: C1) non-verbal communication of object guidance using LED blinking on cart drawers (RCC-LED blinking), C2) verbal communication of object guidance using speech (RCC-Speech), and C3) using a standard crash cart (control).

\subsection{Participants}

We recruited 115 participants, including nurses, child life specialists, and fellows at training levels of 57 4th-year Post-Graduates (PG), 4 5th-year PGs, and 1 6th-year PGs from hospitals and medical schools across the global north.
28 females and 10 males engaged in the RCC-LED blinking condition. 
33 females and 6 males engaged in the RCC-Speech condition. 
26 females and 12 males engaged in the control condition.
We conducted four simulation sessions per condition, with 9-10 participants randomly assigned to each session to balance expertise across teams.
Participants were not compensated for their participation as our research activities were integrated into an existing training event.

\subsection{The Emergency Simulation Room Setup}

The simulation room contained a medical crash cart equipped with a defibrillator, intravenous (IV) supplies, airway equipment, a patient’s vital sign monitor, oxygen tank, a patient bed, and a robotic mannequin controlled by a simulator operator (see Figure \ref{fig:emergecy_room_setup}).
The patient mannequin is a teleoperated platform controlled by a simulator operator that shows lifelike physiological signs such as breathing, bleeding, vital signs, eye blinking, and crying.
Professional actors were hired by a medical school to play the role of a patient’s family member (e.g., parent, sibling, or spouse) to enhance the realism of the resuscitation codes.
The actors often cried, yelled at the clinical team, and impeded care by interrupting the team with questions about the patient’s condition.

\subsection{Study Task}
The local Medical Simulation Center hosting the training event designed the study tasks to mimic real-world patient cases, including high-severity patient cases, and professional actors to portray family members. 
We chose this event-based setting over a traditional lab environment to resemble real-world environments, where we had no control over participants' behaviour, since the deployment of new robots can cause unexpected issues in time-sensitive and critical settings.
Participants engaged in a 15-minute simulated emergency scenario to resuscitate a critically ill and injured 12-month-old boy with ventriculoperitoneal shunt malfunction and subsequent cardiac arrest, working with a RCC. 
We informed participants about the robot’s assistive features, including object search, and demonstrated the robot in experimental conditions before the study session.
The simulation aimed to evaluate primary learning objectives focused on emergency action skills, leadership, mutual support, team decision-making, and family support.

The simulation unfolded in three stages.
During the first five minutes, participants assessed the patient and gathered initial clinical information. 
Between five and seven minutes, they managed the patient’s deterioration. 
Participants were expected to recognize and manage cardiac arrest caused by brain herniation, start high-quality CPR within 30 seconds of cardiac arrest, and address acute increases in intracranial pressure. 
Additionally, participants were asked to show effective team-coordinated decisions to sustain life-saving efforts while addressing legal issues about do-not-resuscitate status, parental authority, and medical decision-making in the absence of instantly available paperwork.

The final stage, lasting from seven to fifteen minutes, focused on responding to cardiac arrest and addressing ethical considerations surrounding do-not-resuscitate decisions.
Furthermore, the simulation required participants to support a member of the patient's family in an appropriate emotional and psychological manner.
Child Life Specialists participants engaged with actors who frequently interrupted medical tasks by asking persistent questions about the patient's condition and requesting detailed explanations of medical information.



\subsection{Wizard-of-Oz Robot Control}

We recruited two medical physician assistants to WoZ control the RCC, allowing healthcare workers to play an active role in demonstrating 'ideal' robot behavior.
Our collaborators hosting the medical training event recruited participants via emails. 
These wizards were well-suited to teleoperate the robot using their medical expertise, which differentiates this study from typical lab-based experimental studies.
We trained the wizards before the study by providing instructions on how to teleoperate the robot and allowing them to practice before study sessions. We instructed the wizards on how to control RCC. They teleoperated the robot based on the emergency scenario. 
They managed either the RCC-LED blinking or RCC-Speech generation to assist participants with object search guidance during the simulation. 
They used a touchscreen tablet with a user interface hidden behind the RCC.
After each session, the wizards handed over the touchscreen to an experimenter to reset the cart to its initial contents.


\subsection{Data Collection \& Analysis}

We collected video data and self-report measures.
We recorded videos of study sessions from one overhead-mounted Insta360 GoPro camera.
We used measurements that positioned HCWs not only as individuals but also as workers within workplace settings.
We administered the NASA Task Load Index (NASA-TLX) \cite{hart1988development} scale, a 7-point Likert scale measure of mental demand, physical demand, temporal demand, effort, and frustration.
We obtained 19 NASA-TLX survey responses in the RCC-Speech condition, 11 in the RCC-LED blinking condition, and 15 in the control condition, resulting in 45 survey responses.
We administered the Attitudes Toward Cooperative Industrial Robots Questionnaire (ACIR-Q) \cite{leichtmann2023new}, a 13-item, 5-point Likert scale measure attitudes toward using robots in the workplace.
We received 14 ACIR-Q responses in the RCC-Speech condition and 9 responses in the RCC-LED blinking condition, resulting in 23 survey responses in total. 
Due to the structure of this training event that closely mirrors actual clinical settings, participants had 10-minute breaks between scenarios, which limited our survey data collection efforts. 


We analyzed the NASA-TLX data using one-way MANOVA.
The assumptions of independence and random selection were satisfied.
We used Shapiro-Wilk tests to check for multivariate normality and Box's M test (p = 0.184) to check for uniformity of covariance matrices.
We used the scatterplot matrices and R² values to examine linearity and showed some violations.
Due to assumption violations, we used Pillai’s Trace for robustness in MANOVA interpretation.
We conducted post-hoc analysis using Bonferroni tests for variables with equal variance ($p > .05$ in Levene’s Test) and the Games-Howell test for variables with unequal variance to determine pairwise differences between experimental conditions.
In the control condition, participants completed the NASA-TLX only because no robot was present.
We analyzed the ACIR-Q data using descriptive statistics.


After all study sessions, we gathered feedback from the wizards for 10 minutes about how participants interacted with the RCC, the challenges wizards faced while controlling the robot, and suggestions for improving the RCC's design, functionality, and verbal- and non-verbal communication. 
We took notes during these sessions and analyzed the feedback notes with thematic analysis \cite{glaser2017discovery}.

\section{Results}

\begin{figure}[t]
    \centering
    \includegraphics[width=0.8\linewidth]{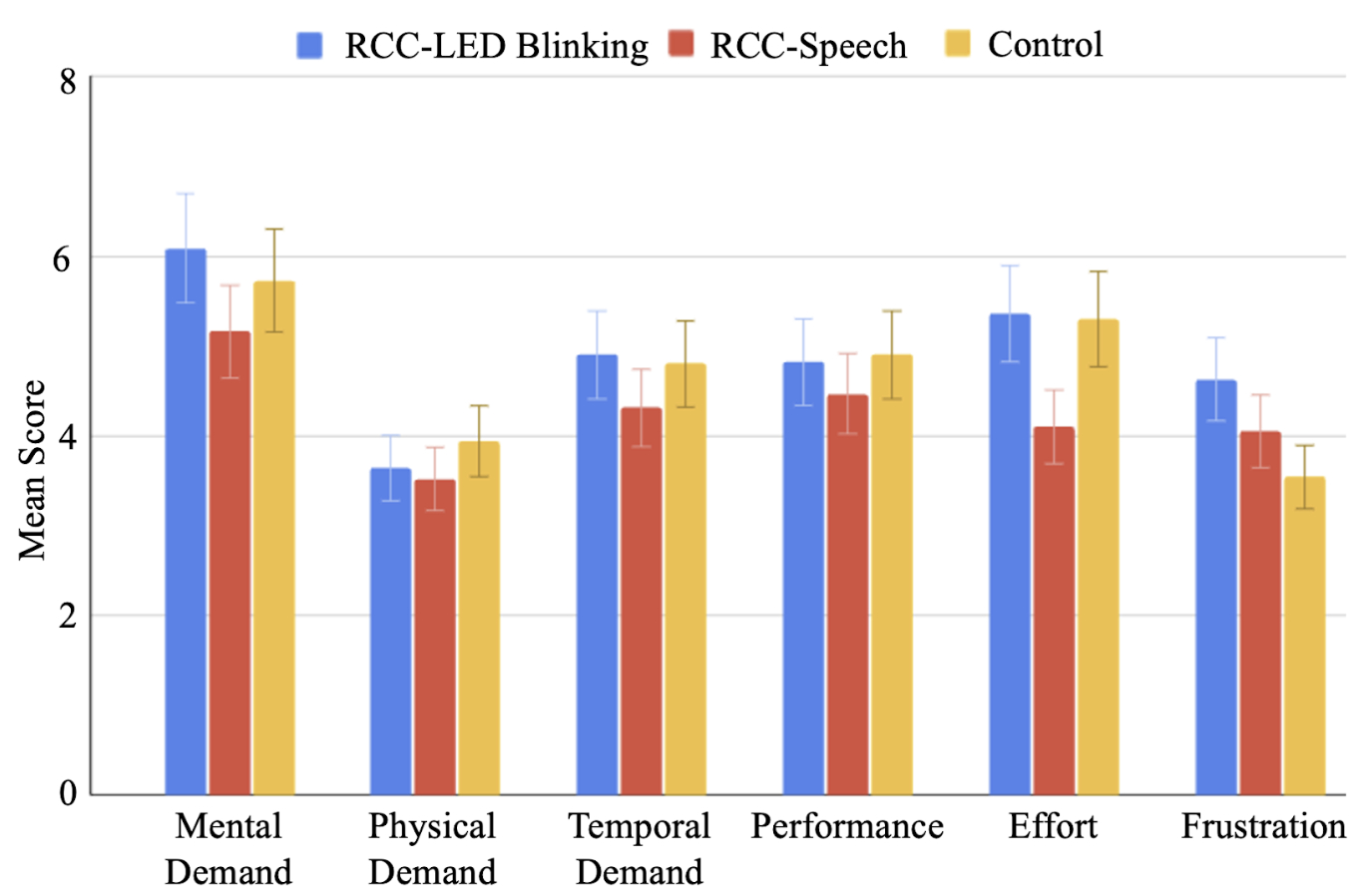} 
    \caption{Participants’ perceived NASA-TLX scores.}
    \label{fig:NASA_scores}
\end{figure}


\textbf{Workload Results:} Figure \ref{fig:NASA_scores} shows the NASA-TLX results. 
According to Wilks’ Lambda test, there was no significant multivariate effect of study conditions on workload ratings, $F(12, 74)=1.690$, $p=.086$; Wilks’ Lambda=$.616$, partial $\eta^2=.215$. In follow-up univariate tests, we discovered a significant effect of study conditions on effort, $F(2, 42)=5.842$, $p=.006$, partial $\eta^2=.218$. The effect of condition on mental demand approached close to significance  $F(2, 42)=3.128$, $p=.054$, partial $\eta^2=.130$. However, we found no significant differences in physical demand, $F(2, 42)=0.365$, $p=.696$, partial $\eta^2=.017$; temporal demand, $F(2, 42)=0.930$, $p=.402$, partial $\eta^2=.042$; performance, $F(2, 42)=0.448$, $p=.642$, partial $\eta^2=.021$; and frustration, $F(2, 42)=1.956$, $p=.154$, partial $\eta^2=.085$.
Post hoc tests showed that effort was significantly lower in the RCC-Speech condition than in the RCC-LED condition ($p=.021$) and the control condition ($p = .019$). Thus, verbal object search required less effort than visual object search and the traditional crash cart.
Additionally, mental demand scored significantly lower in the RCC-Speech condition than in the RCC-LED condition ($p = .042$). 

\begin{figure}[t]
    \centering
    \includegraphics[width=1.0\linewidth]{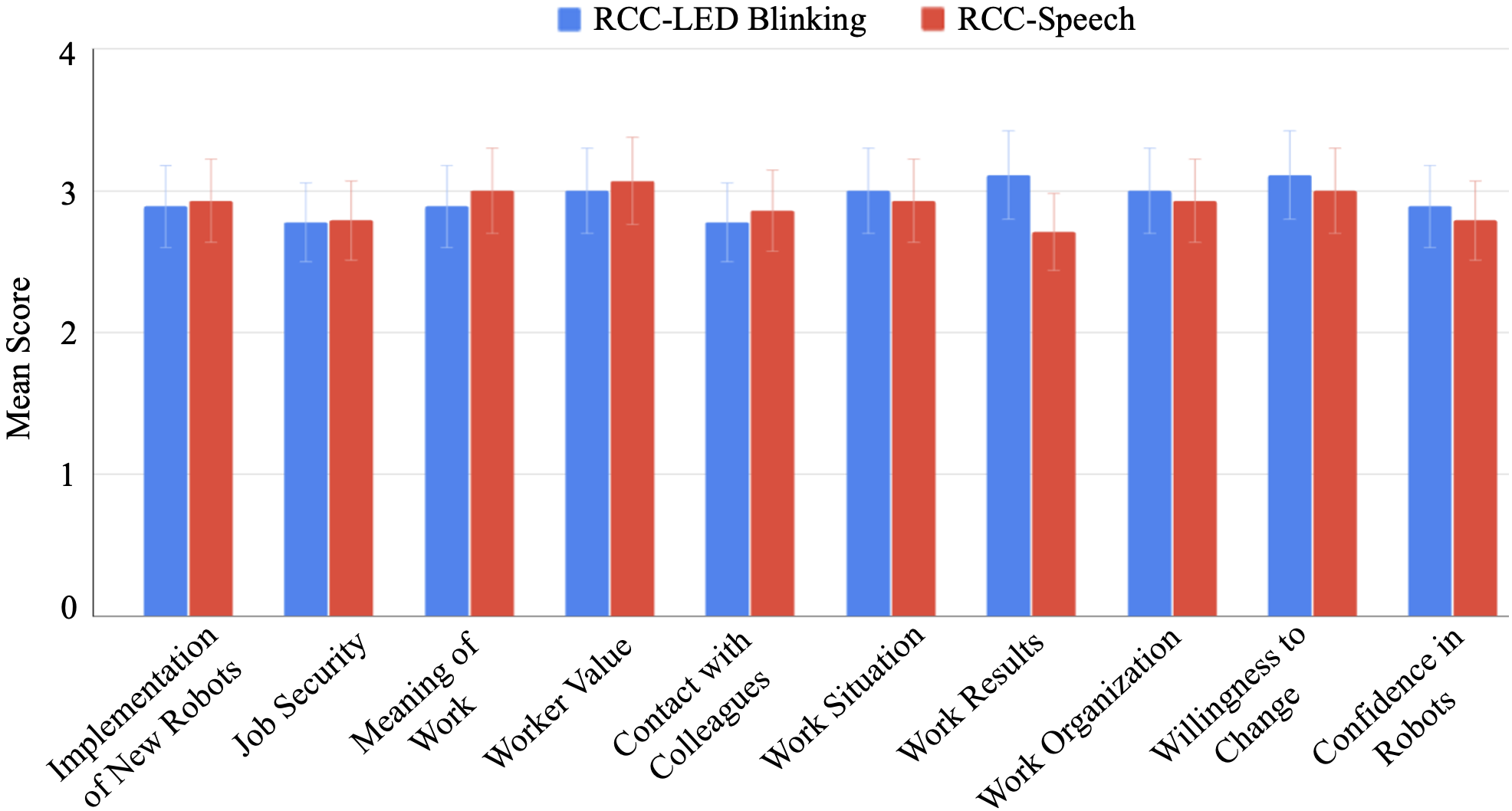} 
    \caption{Participants’ mean ACIR-Q scores.}
    \label{fig:acirq_scores}
\end{figure}


\textbf{ACIR-Q Results:} Figure \ref{fig:acirq_scores} shows the ACIR-Q scores for all conditions. 
The study participants’ acceptance of collaborative robots in the workplace was slightly higher in the RCC-Speech condition (M = 2.93, SD = 1.11) compared to RCC-LED blinking (M = 2.89, SD = 0.78). 
Job security concerns were similar in RCC-LED blinking (M = 2.78, SD = 0.97) and RCC-Speech (M = 2.79, SD = 1.15) conditions.
Participants’ perception of the robot's contribution to making the work meaningful scored slightly higher in RCC-Speech (M = 3.00, SD = 1.06) than in RCC-LED blinking (M = 2.89, SD = 1.17). 
Participants’ perceived worker value was marginally higher in RCC-Speech (M = 3.07, SD = 0.75) than in RCC-LED blinking (M = 3.00, SD = 1.22). 
Concerns about contact with colleagues scored slightly lower in RCC-LED blinking (M = 2.78, SD = 1.20) than in the RCC-Speech condition (M = 2.86, SD = 1.17). 
Participants scored work situation higher in RCC-LED blinking (M = 3.00, SD = 1.41) than in RCC-Speech (M = 2.93, SD = 0.73). 
Robots leading to better work results achieved a higher score in RCC-LED  blinking (M = 3.11, SD = 1.36) compared to RCC-Speech (M = 2.71, SD = 0.77). 
Work organization scored slightly higher in LED blinking (M = 3.00, SD = 1.32) than in RCC-Speech (M = 2.93, SD = 0.75). 
Participants’ willingness to change for a robot followed the same trend, with RCC-LED blinking (M = 3.11, SD = 1.27) scoring higher than RCC-Speech (M = 3.00, SD = 0.90). 
Confidence working with robots was higher in RCC-LED blinking (M = 2.89, SD = 1.17) than in RCC-Speech (M = 2.79, SD = 0.83).


\textbf{Qualitative Wizard Feedback:} The wizards provided critical feedback about the RCC’s use of verbal and non-verbal communication for object search guidance. 
W1 suggested improving the robot's nonverbal cues by placing the LEDs closer to the drawer handles to make them more visible during object search. As stated, ``[Place] lights on drawers and items like an Omnicell.’’
W2 provided useful insights on the robot's verbal prompts to improve clarity, as stated, ``Instead of saying `Epinephrine’ it could say `time for epinephrine.’’’ 
To efficiently reduce participants’ frustration when items are not in the RCC, W2 suggested adding a button to trigger a speech prompt, ``[Add] a button [in UI] for [a speech prompt] `That item is not in this crash cart.’’’ 
W2 also suggested integrating features that would provide the clinicians more control over the RCC’s volume and timing of actions as follows, ``A volume controller on the user interface [and] a stopwatch/timer will help me maintain audible sound in loud environments and track time.”

\section{Discussion}

In this study, we conducted an experimental study by deploying a robot in a high-fidelity clinical simulation environment to understand how robotic verbal and non-verbal cues influence team workload and attitudes toward robots in the workplace.

\noindent\textbf{Impact of Robot Communication on Team Workload:} Our results showed a lower perceived mental demand and effort with a robot that communicates verbally.
Direct verbal prompts helped participants quickly locate items, reducing the need to pause and interpret visual signals.
These findings are consistent with the work of Okamura and Yamada \cite{okamura2020empirical}, which showed that a robot's verbal cues were more reliable than visual cues in human-robot cooperation tasks.
Thus, the design of robot communication strategies for high-stakes action teams requires careful consideration to avoid increasing mental and physical demands \cite{riek2017healthcare, matsumoto2023robot, jamshad2024taking}.

Participants were more frustrated in the experimental condition than in the control condition, although this difference was not significant.
We observed that participants requested items that were missing from the cart, but the robot provided no explicit speech prompt to indicate that those items were not available, which could be one cause of frustration.
Additionally, participants' limited experience with new robotic technology can potentially add complexity to interactions with the robotic cart.
This pattern aligns with research by Hoffman and Breazeal \cite{hoffman2007effects}, who found that robot integration can lead to frustration when the robot does not meet user expectations.
As a result, it is important for robots to communicate their capabilities and limitations to properly establish user expectations. 

\noindent\textbf{Attitudes Toward Collaborative Robotic Systems in the Workplace:} Our study findings indicate that participants were open to the idea of robots in the workplace; however, they have some concerns. Participants rated their attitudes towards robots in the workplace similarly, with the largest difference in workplace results favoring visual cues. This finding was surprising because visual cues result in higher mental demand, temporal demand, and effort compared to standard crash carts. Overall, HCWs showed different preferences for the two modalities when evaluating them as individuals, focusing on efficiency, and as workers, focusing on workplace influence. Their ambivalence suggests that robots may be assessed differently from an individual perspective versus as part of the worker's role within organizational dynamics. This aligns with previous research highlighting the distinction between individual and organizational evaluations of robots \cite{lee2023situating}. Given that most human-robot collaboration studies focus on individual reactions \cite{weiss2021cobots}, more attention is needed on how robots can be evaluated in the context of users' roles as workers and their organizational roles. 

In the future, we plan to increase the robot’s level of autonomy by sensing human interactions in the cart and leveraging closed loop communication to sense conversations to determine what tasks are being performed and supplies needed for those tasks. 
Lastly, we will validate that robot assistance reduces search time of locating relevant supplies to demonstrate the effectiveness of reducing search time and use diverse medical scenarios, including resuscitation for adults. 

\bibliographystyle{IEEEtran}

\begin{thebibliography}{10}
\providecommand{\url}[1]{#1}
\csname url@samestyle\endcsname
\providecommand{\newblock}{\relax}
\providecommand{\bibinfo}[2]{#2}
\providecommand{\BIBentrySTDinterwordspacing}{\spaceskip=0pt\relax}
\providecommand{\BIBentryALTinterwordstretchfactor}{4}
\providecommand{\BIBentryALTinterwordspacing}{\spaceskip=\fontdimen2\font plus
\BIBentryALTinterwordstretchfactor\fontdimen3\font minus \fontdimen4\font\relax}
\providecommand{\BIBforeignlanguage}[2]{{%
\expandafter\ifx\csname l@#1\endcsname\relax
\typeout{** WARNING: IEEEtran.bst: No hyphenation pattern has been}%
\typeout{** loaded for the language `#1'. Using the pattern for}%
\typeout{** the default language instead.}%
\else
\language=\csname l@#1\endcsname
\fi
#2}}
\providecommand{\BIBdecl}{\relax}
\BIBdecl

\bibitem{gray2019workplace}
P.~Gray \emph{et~al.}, ``Workplace-based organizational interventions promoting mental health and happiness among healthcare workers: A realist review,'' \emph{Intern. Journal of Env. Res. and Public Health}, 2019.

\bibitem{bagnasco2013identifying}
A.~Bagnasco \emph{et~al.}, ``Identifying and correcting communication failures among health professionals working in the emergency department,'' \emph{Intern. emergency nursing}, 2013.

\bibitem{o2008professional}
M.~O’Daniel and A.~H. Rosenstein, ``Professional communication and team collaboration,'' \emph{Patient safety and quality: An evidence-based handbook for nurses}, 2008.

\bibitem{Diaz_Dawson_2020}
M.~C. Diaz and K.~Dawson, ``Impact of simulation-based closed-loop communication training on medical errors in a pediatric emergency department,'' \emph{American Journal of Medical Quality}, 2020.

\bibitem{Salik_2023a}
C.~R. Kuehster and C.~D. Hall, ``Simulation: Learning from mistakes while building communication and teamwork,'' \emph{Journal for Nurses in Professional Development}, vol.~26, no.~3, pp. 123--127, 2010.

\bibitem{jacquet2018emergency}
H.~Jacquet, Gabrielle~A \emph{et~al.}, ``The emergency department crash cart: A systematic review and suggested contents,'' \emph{World J. of Emergency Medicine}, vol.~9, no.~2, p.~93, 2018.

\bibitem{makkar2016study}
N.~Makkar and N.~Madaan, ``Study of compliance of crash carts to standards in the emergency of a tertiary care teaching hospital,'' \emph{Int J Res Med Sci}, vol.~4, no.~9, pp. 3968--3976, 2016.

\bibitem{ruggles2010standardizing}
K.~Ruggles, ``Standardizing anesthesia medication drawers using human factors and quality assurance methods,'' \emph{Canadian J. of Anesthesia}, vol.~57, no.~5, p. 490, 2010.

\bibitem{jentsch2016human}
F.~Jentsch, \emph{Human-robot interactions in future military operations}.\hskip 1em plus 0.5em minus 0.4em\relax CRC Press, 2016.

\bibitem{barnes2019human}
M.~Barnes \emph{et~al.}, ``Human-robot interaction design research: from teleoperations to human-agent teaming,'' \emph{Army Research Laboratory}, 2019.

\bibitem{adams2024human}
J.~A. Adams, J.~Scholtz, and A.~Sciarretta, ``Human--robot teaming challenges for the military and first response,'' \emph{Annu. Rev. Control, Robotics, and Autonomous Systems}, vol.~7, 2024.

\bibitem{kruijff2014designing}
K.-K. Kruijff, Geert-Jan~M \emph{et~al.}, ``Designing, developing, and deploying systems to support human--robot teams in disaster response,'' \emph{Advanced Robotics}, vol.~28, no.~23, pp. 1547--1570, 2014.

\bibitem{bordini2009formal}
R.~H. Bordini, M.~Fisher, and M.~Sierhuis, ``Formal verification of human-robot teamwork,'' in \emph{Proc. 4th / international conference on Human robot interaction}, 2009, pp. 267--268.

\bibitem{gervits2018shared}
F.~Gervits, T.~W. Fong, and M.~Scheutz, ``Shared mental models to support distributed human-robot teaming in space,'' in \emph{2018 aiaa space and astronautics forum and exposition}, 2018, p. 5340.

\bibitem{taylor2019coordinating}
A.~Taylor, H.~R. Lee, A.~Kubota \emph{et~al.}, ``Coordinating clinical teams: Using robots to empower nurses to stop the line,'' \emph{Proc. on Human-Computer Interaction}, vol.~3, no. CSCW, pp. 1--30, 2019.

\bibitem{taylor2024towards}
A.~Taylor, T.~Tanjim, H.~Cao, and H.~R. Lee, ``Towards collaborative crash cart robots that support clinical teamwork,'' in \emph{Proceedings of the 2024 ACM/IEEE International Conference on Human-Robot Interaction}, 2024, pp. 715--724.

\bibitem{taylor2025rapidly}
A.~Taylor, T.~Tanjim, M.~J. Sack, M.~Hirsch, K.~Cheng, K.~Ching, J.~S. George, T.~Roumen, M.~F. Jung, and H.~R. Lee, ``Rapidly built medical crash cart! lessons learned and impacts on high-stakes team collaboration in the emergency room,'' in \emph{2025 20th ACM/IEEE International Conference on Human-Robot Interaction (HRI)}.\hskip 1em plus 0.5em minus 0.4em\relax IEEE, 2025, pp. 501--510.

\bibitem{rivera2010interruptions}
A.~Rivera-Rodriguez and B.-T. Karsh, ``Interruptions and distractions in healthcare: review and reappraisal,'' \emph{BMJ Quality \& Safety}, 2010.

\bibitem{sutherland2023fatigue}
C.~Sutherland, A.~Smallwood \emph{et~al.}, ``Fatigue and its impact on performance and health,'' \emph{British J. of Hospital Medicine}, vol.~84, no.~2, pp. 1--8, 2023.

\bibitem{nourbakhsh2005human}
I.~R. Nourbakhsh, K.~Sycara \emph{et~al.}, ``Human-robot teaming for search and rescue,'' \emph{Pervasive Computing}, vol.~4, no.~1, pp. 72--79, 2005.

\bibitem{chen2014human}
J.~Y. Chen and M.~J. Barnes, ``Human--agent teaming for multirobot control: A review of human factors issues,'' \emph{IEEE Trans. on Human-Machine Systems}, vol.~44, no.~1, pp. 13--29, 2014.

\bibitem{ma2022metrics}
L.~M. Ma, M.~Ijtsma, K.~M. Feigh \emph{et~al.}, ``Metrics for human-robot team design: A teamwork perspective on evaluation of human-robot teams,'' \emph{ACM Trans. on Human-Robot Inter. (THRI)}, 2022.

\bibitem{haripriyan2024human}
A.~Haripriyan, R.~Jamshad \emph{et~al.}, ``Human-robot action teams: A behavioral analysis of team dynamics,'' in \emph{Intern. Conf. on Robot and Human Interactive Communication (ROMAN)}, 2024, pp. 1443--1448.

\bibitem{jamshad2024taking}
A.~Jamshad, R. abd~Haripriyan \emph{et~al.}, ``Taking initiative in human-robot action teams: How proactive robot behaviors affect teamwork,'' in \emph{Intern. Conf. on Human-Robot Interaction}, 2024, pp. 559--562.

\bibitem{su2023recent}
S.~Hang, W.~Qi \emph{et~al.}, ``Recent advancements in multimodal human--robot interaction,'' \emph{Front. Neurorobot.}, vol.~17, p. 1084000, 2023.

\bibitem{Bonarini_2020}
A.~Bonarini, ``Communication in human-robot interaction,'' \emph{Current Robotics Reports}, vol.~1, no.~4, p. 279–285, 2020.

\bibitem{szafir2015communicating}
D.~Szafir, B.~Mutlu, and T.~Fong, ``Communicating directionality in flying robots,'' in \emph{Proc. of Intern. Conf. on Human-Robot Inter.}, 2015.

\bibitem{kaszuba2023speech}
S.~Kaszuba, S.~R. Sabbella \emph{et~al.}, ``Speech act classification in collaborative robotics,'' in \emph{2023 32nd Intern. Conf. on Robot and Human Interactive Communication (RO-MAN)}, 2023, pp. 2169--2175.

\bibitem{kamboj2022examining}
A.~Kamboj, T.~Ji, and K.~Driggs-Campbell, ``Examining audio communication mechanisms for supervising fleets of agricultural robots,'' in \emph{2022 31st Intern. Conf. on Robot and Human Interactive Communication (RO-MAN)}, 2022, pp. 293--300.

\bibitem{tang2024assisting}
F.~Tang, C.~Zheng \emph{et~al.}, ``Assisting group discussions using desktop robot haru,'' in \emph{Intern. Conf. on Robotics and Automation}, 2024.

\bibitem{breazeal2005effects}
C.~Breazeal, C.~D. Kidd \emph{et~al.}, ``Effects of nonverbal communication on efficiency and robustness in human-robot teamwork,'' in \emph{Intern. conf. on intelligent robots and systems}, 2005, pp. 708--713.

\bibitem{brown2024trash}
B.~Brown, F.~Bu, I.~Mandel \emph{et~al.}, ``Trash in motion: Emergent interactions with a robotic trashcan,'' in \emph{Proc. 2024 CHI Conf. on Human Factors in Computing Systems}, 2024, pp. 1--17.

\bibitem{lee2023situating}
H.~R. Lee, X.~Tan, W.~Zhang \emph{et~al.}, ``Situating robots in the organizational dynamics of the gas energy industry: A collaborative design study,'' in \emph{Proc. IEEE Intern. Conf. on Robot and Human Interactive Communication (ROMAN}, 2023, pp. 1096--1101.

\bibitem{alami2006safe}
R.~Alami, A.~Albu-Schäffer, A.~Bicchi \emph{et~al.}, ``Safe and dependable physical human-robot interaction in anthropic domains: State of the art and challenges,'' in \emph{Proc. IEEE/RSJ Intern. Conf. on Intelligent Robots and Systems (IROS)}, 2006, pp. 1--16.

\bibitem{raimondo2022trailblazing}
F.~R. Raimondo, A.~T. Wolff \emph{et~al.}, ``Trailblazing roblox virtual synthetic testbed development for human-robot teaming studies,'' in \emph{Proc. Human Factors and Ergonomics Society Annual Meeting}, 2022.

\bibitem{wong2021remote}
M.~Wong, A.~Ezenyilimba \emph{et~al.}, ``A remote synthetic testbed for human-robot teaming: an iterative design process,'' in \emph{Proc. human factors and ergonomics society annual meeting}, 2021.

\bibitem{heinold2024advaned}
E.~Heinold, P.~Rosen, and S.~Wischniewski, ``Advaned robots in healthcare and their impact on the health and safety of medical workers,'' in \emph{2024 33rd Intern. Conf. on Robot and Human Interactive Communication (ROMAN)}, 2024, pp. 258--263.

\bibitem{hart1988development}
S.~Hart, ``Development of nasa-tlx: Results of empirical and theoretical research,'' \emph{Human mental workload/Elsevier}, 1988.

\bibitem{leichtmann2023new}
B.~Leichtmann, J.~Hartung, O.~Wilhelm \emph{et~al.}, ``New short scale to measure workers’ attitudes toward the implementation of cooperative robots in industrial work settings: Instrument development and exploration of attitude structure,'' \emph{Int. J. Social Robotics}, vol.~15, no.~6, pp. 909--930, 2023.

\bibitem{glaser2017discovery}
B.~Glaser and A.~Strauss, \emph{Discovery of grounded theory: Strategies for qualitative research}.\hskip 1em plus 0.5em minus 0.4em\relax Routledge, 2017.

\bibitem{okamura2020empirical}
K.~Okamura and S.~Yamada, ``Empirical evaluations of framework for adaptive trust calibration in human-ai cooperation,'' \emph{Access}, 2020.

\bibitem{riek2017healthcare}
L.~D. Riek, ``Healthcare robotics,'' \emph{Communications of the}, 2017.

\bibitem{matsumoto2023robot}
S.~Matsumoto, P.~Ghosh \emph{et~al.}, ``Robot, uninterrupted: Telemedical robots to mitigate care disruption,'' in \emph{Intern. Conf. on Human-Robot Inter.}, 2023.

\bibitem{hoffman2007effects}
G.~Hoffman and C.~Breazeal, ``Effects of anticipatory action on human-robot teamwork efficiency, fluency, and perception of team,'' in \emph{Intern. Conf. on Human-robot Inter.}, 2007.

\bibitem{weiss2021cobots}
A.~Weiss, A.-K. Wortmeier, and B.~Kubicek, ``Cobots in industry 4.0: A roadmap for future practice studies on human--robot collaboration,'' \emph{IEEE Trans. Human-Machine Systems}, vol.~51, no.~4, pp. 335--345, 2021.

\end{thebibliography}

\end{document}